\documentclass[letterpaper]{article} 
\usepackage[draft]{aaai25}  
\usepackage{times}  
\usepackage{helvet}  
\usepackage{courier}  
\usepackage[hyphens]{url}  
\usepackage{graphicx} 
\usepackage{ctable}
\usepackage{natbib}  
\usepackage{caption} 
\usepackage{algorithm}

\usepackage{newfloat}
\usepackage{listings}
\DeclareCaptionStyle{ruled}{labelfont=normalfont,labelsep=colon,strut=off} 
\floatstyle{ruled}
\newfloat{listing}{tb}{lst}{}
\floatname{listing}{Listing}
\usepackage{algpseudocode}
\usepackage{multirow}
\usepackage{multicol}
\usepackage{xcolor}
\usepackage{amsmath}

\title{Resource-Constrained Heuristic for Max-SAT}
\author{
    Brian Matejek\textsuperscript{\rm 1},
    Daniel Elenius\textsuperscript{\rm 1},
    Cale Gentry\textsuperscript{\rm 2}, 
    David Stoker\textsuperscript{\rm 2}, 
    Adam Cobb\textsuperscript{\rm 1}
}
\affiliations{
    \textsuperscript{\rm 1}Computer Science Laboratory, SRI International\\
    \textsuperscript{\rm 2}Advanced Technology and Systems Division, SRI International\\    
}

\begin{document}

\maketitle

\begin{abstract}
We propose a resource-constrained heuristic for instances of Max-SAT that iteratively decomposes a larger problem into smaller subcomponents that can be solved by optimized solvers and hardware. The unconstrained outer loop maintains the state space of a given problem and selects a subset of the SAT variables for optimization independent of previous calls. The resource-constrained inner loop maximizes the number of satisfiable clauses in the “sub-SAT” problem. Our outer loop is agnostic to the mechanisms of the inner loop, allowing for the use of traditional solvers for the optimization step. However, we can also transform the selected “sub-SAT” problem into a quadratic unconstrained binary optimization (QUBO) one and use specialized hardware for optimization. In contrast to existing solutions that convert a SAT instance into a QUBO one before decomposition, we choose a subset of the SAT variables before QUBO optimization. We analyze a set of variable selection methods, including a novel graph-based method that exploits the structure of a given SAT instance. The number of QUBO variables needed to encode a (sub-)SAT problem varies, so we additionally learn a model that predicts the size of sub-SAT problems that will fit a fixed-size QUBO solver. We empirically demonstrate our results on a set of randomly generated Max-SAT instances as well as real world examples from the Max-SAT evaluation benchmarks and outperform existing QUBO decomposer solutions.
\end{abstract}

\section{Introduction}

Max-SAT is an NP-Hard optimization problem that asks for the maximum number of satisfiable clauses of a Boolean formula written in conjunctive normal form (CNF)~\cite{krentel1986complexity}. For an $N$ variable problem, this requires searching  over the space of $2^N$ possible truth assignments. This generalization of the satisfiability problem (SAT) has significant applications in various sectors~\cite{asin2014curriculum,fu2006solving}. Although there are a number of algorithms and commercial tools for exactly solving Max-SAT~\cite{argelich2006exact,de2008z3,xing2005maxsolver}, these techniques cannot scale to arbitrarily large problem instances. Therefore, a significant amount of research has focused on incomplete (or anytime) solvers that provide a solution without guaranteeing correctness. These algorithms generally fall into two categories: local heuristics and global approximation approaches. In this paper we focus on a local neighborhood search heuristic (LNS) that continually optimizes small sub-problems of the full SAT instance. 

We propose a resource-constrained heuristic for Max-SAT problems that operates in a two-tier framework. The outer-loop iteratively selects a sub-selection of SAT variables and extracts a partial CNF formula. An inner-optimizer proposes a variable assignment for the sub-problem. A composer merges this solution with the global one to reduce the energy of the current, global, variable assignment. Our method is agnostic to the mechanisms of both the selector and optimizer, allowing us to use specialized hardware as our optimizer. Our approach contrasts with existing solutions that first reformulate a SAT instance into a QUBO problem before decomposition, optimization, and composition. We devise a novel graph-based selection method for choosing small sub-SAT instances. Since our framework is agnostic to the inner-optimizer, we show results using exact and anytime optimizers. Additionally, we can convert our sub-SAT instances into QUBO problems  and use existing solvers with constrained, but optimized, hardware. Selecting variables in the SAT space before producing a QUBO outperforms existing strategies that convert a SAT problem into a QUBO one before decomposition. We demonstrate our results on a series of random Max-SAT instances, as well as examples from the Max-SAT 2016 challenge benchmark. 
\section{Background and Notation}

A propositional logic formula or Boolean expression in conjunctive normal form (CNF) contains a series of $L$ clauses each comprised of a (sub)set of $N$ literals. In each clause, we define each variable as either a positive or negative literal, depending on the inclusion of a negation sign for a given literal. Each clause is a disjunction of the variables and a given formula is a conjunction of all clauses. A CNF is \textit{satisfiable} if some assignment of \textit{true} or \textit{false} for each variable satisfies each clause. Max-SAT seeks to find an assignment that satisfies the maximum number of clauses over all possible assignments. For both $k$-SAT and Max-$k$SAT the problems constrain the number of variables per clause $\leq k$. We use the following notation for a given Max-SAT instance with $N$ variables and $L$ clauses. Clauses and variables are uniquely labeled $c_1, ... ,c_l, ... ,c_L$ and $x_1, ... , x_i, ... , x_N$, respectively. A clause $c_l$ contains $k \geq 1$ positive or negative literals $x_i, x_j, ... $ with a negation sign, $\neg$, indicating a negative literal. For this work, we have focused on $3$-SAT instances, although the methodology extends to larger values of $k$. At a given timestamp, the \textit{state} of a SAT instance refers to the current assignment of \textit{true} or \textit{false} values to each variable. The \textit{energy} of a given state indicates the number of unsatisfied clauses under that assignment. A state with zero energy satisfied the $k$-SAT instance. The goal of Max-SAT is to find the state corresponding to the lowest possible energy given a Boolean formula.

Complete, or exact, solvers for Max-SAT guarantee a correct solution. Traditionally, Max-SAT incomplete solvers have fallen into two categories: local heuristics and global approximation algorithms. Local heuristics such as Walk-SAT~\cite{selman1993local} have the advantage of being quick exploration mechanisms but may get caught in a local minima and thus have no guarantees on correctness. Various strategies such as simulated annealing and random restarts can allow these local algorithms to jump out of local minima and better explore the problem space~\cite{hoos2000local,spears1993simulated}. The G-SAT algorithm iteratively flips the truth assignment of the literal that reduces the energy most of the SAT instance~\cite{selman1992new}. Approximation algorithms often convert the Max-SAT instance into an integer linear program and use linear programming to solve a relaxed version of the problem~\cite{karloff19977,sinjorgo2023solving}, requiring more resources but sometimes guaranteeing proximity to the optimal solution.

As a middle ground, partition strategies typically separate clauses into groups at the start of the solving process~\cite{morgado2013iterative}. These strategies then hierarchically merge together a small subset of partitions to gradually construct a globally optimal solution that satisfies the maximum number of clauses. Orvalho~\textit{et al.} propose UpMax, a method that decouples the partitioning process from the MaxSAT solver allowing a user to manually partition soft clauses based on domain knowledge of the problem~\cite{orvalho2023}. Our method most closely resembles existing work in ``Large Neighborhood Search''~\cite{shaw1998using,hickey2022large}. These solutions repeatedly take a small sub-section of a larger problem for optimization, typically calling an exact solver. We differ from these methods by allowing for arbitrary inner-optimizers, including ones that transform the selected neighborhood into a QUBO optimization problem. 

Tangentially, exact SAT solvers for quantum~\cite{alasow2022quantum} and analog computers offer promise of fast solutions on small instances that can fit on the hardware~\cite{molnar2018continuous}. However, these methods cannot scale to the larger problem instances until the underlying circuitry does. As a solution, D-Wave proposes \texttt{Qbsolv} which first converts a SAT instance into a QUBO matrix optimization formulation~\cite{booth2017partitioning}. An outer-loop algorithm selects a sub-matrix from the full QUBO (a sub-QUBO), and optimizes the sub-QUBO using the available hardware. It then composes the solution on the sub-QUBO back into the full QUBO solution space. Converting a $k$-SAT problem into a QUBO one requires auxiliary variables for $k > 2$ (i.e., the number of QUBO variables will exceed the number of SAT literals). In the worst case, a $3$-SAT conversion will generate $N + L$ QUBO variables where $N$ and $L$ are the number of SAT literals and clauses, respectively~\cite{chancellor2016direct}. We directly compare to this sub-QUBO approach and show that sub-selection of the SAT variables prior to the QUBO conversion and inner-optimization step leads to better performance.
\section{Methodology}

We propose a two component heuristic for optimizing Max-SAT instances, where the optimization routine is constrained in the number of variables it can handle. In particular, our algorithm contains an outer-loop and an inner-optimizer. The inner-optimizer is constrained by the number of variables over which it can operate. Therefore the outer-loop is limited in the size of the sub-SAT it can pass to the inner-optimizer. Note, by maximum sub-SAT size, we constrain only the number of variables sent, $M$, not the number of clauses.

During optimization, we find an assignment that minimizes the number of unsatisfied clauses in the sub-SAT problem, independent of the rest of the full problem space. During \textbf{decomposition}, we enforce any constraints imposed by the inner-optimizer such as QUBO size or sub-SAT size. The outer-loop is agnostic to the internal mechanisms of the \textbf{inner-optimizer}, allowing us to use exact, anytime, or QUBO solvers as the inner-optimizer. After the inner-optimizer concludes, a \textbf{composer} updates the global solution with the optimized variables. We provide pseudocode for various components of our algorithm in the supplementary material.

\subsection{Decomposer}

We propose four different methods for selecting $M$ variables from a 3-SAT instance: random, energy-based, softmax, and graph-based. In each instance, the method takes as input the current state of the Max-SAT problem and the number of SAT variables, $M$, to return. 

\paragraph{Random Selector.} 
The random selector returns $M$ random SAT variables regardless of the state or structure of the CNF instance. This simple selector allows one to explore the state space without existing bias, which can help break out of local minima.

\paragraph{Energy Selector.} 
The energy-based selector identifies the variables that produce the largest delta between the two possible assignments, assuming all other variables remain frozen. That is, we look at the current energy of an assignment versus the energy when switching each variable independently. We can order these differences to create a ranking of the variables that would improve the energy the most if switched. We select the $M$ variables that provide the most improvement in energy. These variables, when switched independently, provided the largest increase in satisfied clauses. Note, the energy selector is similar to the inner loop of GSAT~\cite{selman1992new}. However, as opposed to selecting one variable for flipping, we select $M$ variables for optimization. This is also the approach that is used by \texttt{Qbsolv}.

\subsubsection{Softmax Selector.}

The softmax selector is similar to prior work of Hickey~\textit{et al.} on the Large Neighborhood Search (LNS) method~\cite{hickey2022large}. The selector first calculates the potential changes in energy when flipping each variable. We then normalize the potential energy changes into an array $e$ using the softmax function. We then sample $M$ variables without replacement according to these softmax probabilities.

\begin{figure}[t!]
    \centering
    \includegraphics[width=0.85\linewidth]{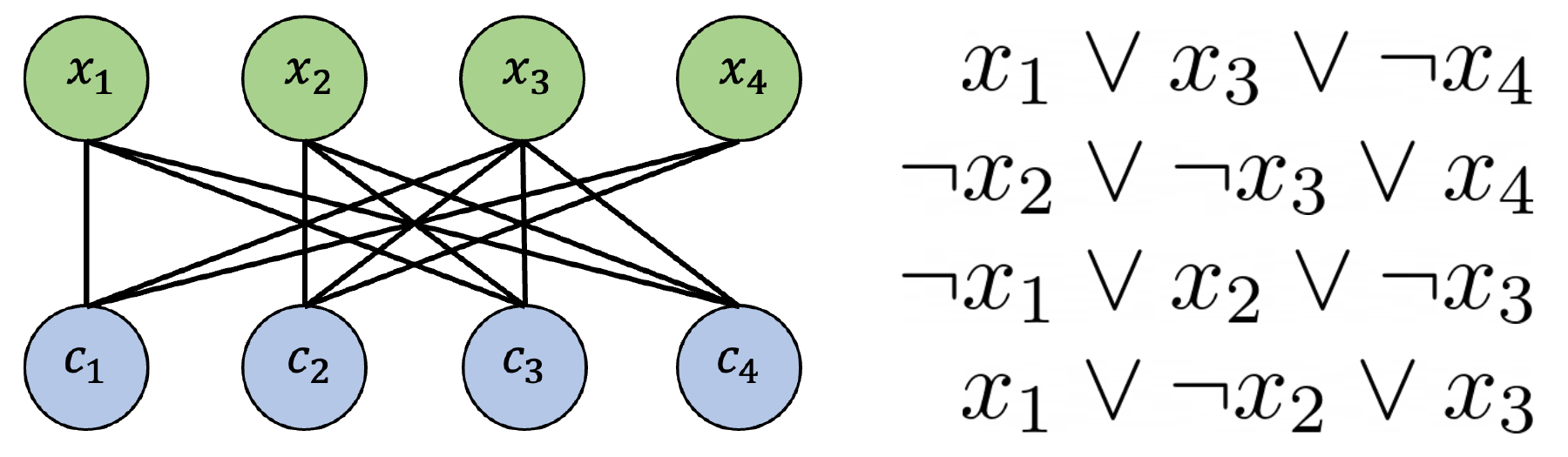}
    \caption{We construct a weighted bipartite graph from any k-SAT instance. Each of the $N$ variables and $L$ clauses receives a single node in the graph. We add an edge between nodes $x_i$ and $c_j$ if clause $c_j$ contains variable $x_i$.}
    \label{figure:graph-representation}
\end{figure}

\paragraph{Graph Selector.} Our graph-based selector attempts to exploit the internal structure of the CNF instance by finding sets of clauses that have a high overlap of variables. Furthermore, we identify the variable/clause pairs that have a high degree of unsatisfiability (i.e., the unsatisfied clauses in the current state). We model any $k$-SAT instance as a weighted bipartite graph (Figure~\ref{figure:graph-representation}). Our graphical model matches the Clause Variable-Incidence Graph common in the literature~\cite{ansotegui2012community}, with some changes to the edge weighting scheme.

We construct a graph with $L + N$ nodes with one node for each clause and variable. We do not differentiate between positive and negative literals of the same variable. For every variable $x_i$ in clause $c_j$ we add an edge between the corresponding nodes labeled $x_i$ and $c_j$. The number of edges has an upper bound of $kL$, with fewer edges if some clauses have fewer than $k$ literals. For simplicity, we define the set of vertices in the graph as $V \in v_1 ... v_{L + N}$, and the set of edges as $E \in (v_i, v_j)$.

We apply weights to the edges based on the state of the $k$-SAT instance. We look at every clause and determine the number of unsatisfied literals (i.e., a positive literal where the variable is \textit{false}, or a negative literal where the variable is \textit{true}). If there are $n$ unsatisfied literals for a clause $c_l$, we assign an edge weight of $f(n)$ to all edges adjacent to the node $c_l$. $f(x) \to \mathcal{R}$ is a real-valued function that takes an integer and produces an edge weight. We only consider functions $f$ that are monotonically increasing so that clauses with more unsatisfied literals have adjacent edges with higher values. 

After graph construction, we run an iterative algorithm to find a subset of variable nodes and clauses that have a high average weighted degree. Our goal is to identify subgraphs (clusters of nodes and clauses) where there is a high-level of overlap between the nodes in clauses and the clauses are unsatisfiable. We begin by selecting $n$ variable nodes and add them to a set of ``in'' nodes. The ``in'' nodes, and the edges that connect them, will comprise our current subgraph. Note, our first subgraph will not contain any edges since we select only variables nodes and our graph is bipartite. We construct two sets: $\mathcal{I}$ and $\mathcal{O}$, representing the nodes that are ``in'' the current subgraph and those that are ``out'' of the current subgraph. A vertex is either in $\mathcal{I}$ or $\mathcal{O}$, and $\mathcal{I} \cup \mathcal{O} = V$ and $\mathcal{I} \cap \mathcal{O} = \emptyset$. For each vertex $v_i$, we define the ``connectivity'' function:
\begin{equation}
    c(v_i) = \sum_{v_j \in \mathcal{I}, (v_i, v_j) \in E} w_{ij}
\end{equation}
The connectivity function measures the sum of edge weights between the vertex $v_i$ and its neighbors in $\mathcal{I}$.

During each iteration of the algorithm, we select the vertex $v_i \in \mathcal{O}$ with the maximum $c(v_i)$ value and the vertex $v_j \in \mathcal{I}$ with the minimum $c(v_j)$ value. We swap $v_i$ and $v_j$ into the opposite set and update all $c$ values for the other vertices. At the first iteration, we will remove a vertex corresponding to a variable with one corresponding to a clause. However, we want to maintain the number of variable vertices in the set $\mathcal{I}$ for our sub-SAT selector. Thus, if the number of variable nodes ever drops below our target size, we add the variable vertex in $\mathcal{O}$ with the highest $c$ value to $\mathcal{I}$. Conversely, if the number of variable nodes exceeds our target size, we remove the variable vertex with the lowest $c$ value in $\mathcal{I}$.

\subsection{Clause Pruning}

At this point, we have selected the \textit{dynamic} ($M$) and \textit{frozen} ($N-M$) variables. Any clause that does not contain any \textit{dynamic} variables can be removed from the sub-SAT instance since its state cannot change. Furthermore, since the assignment of the \textit{frozen} variables cannot change in the inner-optimizer, certain clauses may remain satisfied despite changes to the dynamic variables. For example, if $c_l$ is $x_i \lor \lnot x_j \lor x_k$, and $x_j$ is frozen as \textit{false}, $c_l$ will remain satisfied regardless of changes to $x_i$ and $x_k$. Therefore, we can exclude this clause from the inner loop. We can also use the \textit{frozen} variables to reduce the number of literals per clause, which is significant when we introduce inner QUBO optimizers, since clauses with $k \leq 2$ can be directly represented in QUBO problems without the need for additional variables. This is a significant strength of our approach when reducing to QUBO after the decomposer.

\subsection{Inner-Optimizer}

The outer-loop mechanism for selecting a subset of SAT variables is only constrained by the number of variables that can be operated on by the inner-optimizer. For this work, we have focused on three inner-optimizer algorithms, including traditional SAT solvers and QUBO optimizers. The QUBO optimizers come with the additional challenge of the translation to QUBO. Therefore, for QUBO, the constraint on the number of variables corresponds to the QUBO size $Q$, which presents itself as a challenge, since the decomposer provides an $M$. We overcome this using a cheap linear regression model in a novel way.

\paragraph{SAT Solvers.}
We use two optimizers in this work: z3, an exact solver~\cite{de2008z3}, and Walk-SAT, a local heuristic~\cite{selman1993local}. 

\paragraph{QUBO Optimizers.}
The motivation for converting to QUBO problems comes from significant research focused on specialized hardware specifically designed to QUBO problems~\cite{booth2017partitioning,date2019efficiently,kalinin2023analog}. Such hardware comes with physical constraints on the number of variables can fit on the hardware. However, their performance power can still be leveraged if we can ensure the problems are small enough. The challenge of incorporating a QUBO optimizer within our optimization loop comes from the need to translate from the $M$ variable sub-SAT problem to a $Q$ variable QUBO problem. The direct mapping from a 3-SAT problem formulation to a QUBO problem uses the following construction to define the energy $\Phi(x)$:
$$\sum_{i=1}^L (x_{i1}+x_{i2}+x_{i3}-x_{i1} x_{i2}-x_{i1} x_{i3}-x_{i2} x_{i3}+x_{i1} x_{i2} x_{i3})$$
where $L$ is the number of clauses. This pseudo-Boolean function contains a cubic term that we can remove by adding an extra variable $w_i \in \{0, 1\}$ per clause such that $x_{i1} x_{i2} x_{i3} = \max_{w_i} w_i (x_{i1}+x_{i2}+x_{i3}-2)$. Thus, we define the energy $\Phi(x)$:
$$\sum_{i = 1}^L (1 + w_i)(x_{i1} + x_{i2} + x_{i3}) - x_{i1}x_{i2} - x_{i1}x_{i3} - x_{i2}x_{i3} - 2w_i$$
If $x_{i1}, x_{i2},$ and $x_{i3}$ all evaluate to \textit{true}, $w_i = 1$; if two of the three literals evaluate to \textit{true}, $w_i$ can be 0 or 1; if one of the three literals evaluate to \textit{true}, $w_i = 0$; if none of the literals evaluate to \textit{true}, $w_i = 0$. One can check that the expression for each clause thus evaluates to 1 if the clause is satisfied, and 0 otherwise. As a result, an $N$ variable, $L$ clause 3-SAT problem would generate a QUBO problem with $N + L$ variables QUBO problem using this default formulation. For converting clauses with $k < 3$, we can remove the auxiliary cubic terms. For clauses with $k > 3$, we can first convert to $3$-SAT before QUBO conversion.

A significant amount of optimization has focused on reducing the number of auxilary variables needed when converting a SAT problem into a QUBO one~\cite{zaman2021pyqubo}, and the methods for generating those variables~\cite{nusslein2023black,chancellor2016direct}. For our work, we focus on methods from Chancellor~\textit{et al.}~\cite{chancellor2016direct} and N{\"u}\ss lein~\textit{et al.}~\cite{nusslein2023black}. Since we call the QUBO conversion method for each iteration of the outer-loop, we prioritized speed over a reduction in auxiliary variables. This speed tradeoff is less critical in existing strategies that convert a full SAT instance into a QUBO formulation exactly once. 

\textit{Learned Variable Mapping.} Going back to the challenge of translating from the $M$ variable sub-SAT problem to the $Q$ variable QUBO problem, we need a way to impose the $Q$ constraint on the outer-loop. Our interesting solution is to use a simple linear regression model that takes five inputs to generate a candidate $M$. These inputs include the final maximum QUBO size, $Q$, the maximum number of literals in any clause, the total number of literals in the CNF, $N$, and $L$. To account for when the linear regression model predicts an $M$ that leads to a $Q$ too large for the optimizer, we include a binary search as a backup. 
Therefore, we perform a binary search as well to identify the optimal number of SAT variables $M$ given the QUBO matrix constraint $Q$. We first use our learned predictor to determine a preliminary $M$. If converting the returned sub-SAT instance results in a QUBO matrix that exceeds the hardware constraints, we perform a binary search to identify the optimal $M^*$ that produces a QUBO matrix within the hardware constraints. If we could have fit a larger QUBO matrix on the hardware, we increase the value of $M$ for the next call to the optimizer. If we exceed the value of $Q$, we reduce the value of $M$ before calling the inner-optimizer. The function from $M \to Q$ is stochastic and depends on the variables returned by the selector. Thus, we continually check to confirm that the QUBO produced with a sub-SAT selection of $M$ variables fit within the constraints of the optimizer.

\textit{Tabu Search.} The specific algorithm that we use to optimize the QUBO problems in this paper is Tabu search \cite{palubeckis2004multistart}. Tabu search iteratively chooses variables to flip. Once the algorithm chooses a variable for flipping, it stores the variable for a set number of iterations in the tabu list. Variables on this list cannot change for a set number of iterations. There are many variants of Tabu search; for our work we use the implementation provided by D-Wave's \texttt{Qbsolv}~\cite{booth2017partitioning}. 
Finally, we convert from the solution in the QUBO space to the SAT space. 

\subsection{Composer}

During the composition step, we update the assignment of the \textit{dynamic} variables based on the output from our inner-optimizer. By decoupling the sub-SAT selection method from the solver, we avoid the issues with rectifying multiple disjoint solutions into a coherent best choice. 

\subsection{Early Stopping}

After each iteration of the outer-loop, we calculate the energy of the current state of the solution. If the energy doesn't improve after $conv$ number of iterations, we terminate the process. Otherwise, we terminate after $max\_iters$ number of calls to the inner-optimizer. We find that some selectors, such as Energy and Softmax, require fewer iterations to converge to a local minima. Other selectors, such as the Random and Graph approaches, generally reach better minima but only after more iterations. 
\section{Experimental Setup}

\subsubsection{Datasets.}

We demonstrate our results on a range of SAT instances. We make our data and code publicly available.\footnote{See supplementary materials.}
We construct a set of random datasets with 100, 200, and 500 variables and clause to variable ratios of 4.00, 4.25, and 4.50. These three ratios correspond to satisfiable, a mix, and unsatisfiable random instances. We use the \texttt{randkcnf} formula from the \texttt{CNFgen} toolkit to generate these random 3-SAT instances~\cite{lauria2017cnfgen}. 
We consider a subselection of the handcrafted Max-Cut instances from the Max-SAT 2016 challenge (unweighted benchmark)\footnote{http://maxsat.ia.udl.cat/benchmarks/}. These 1,417 CNF instances range from 150 and 220 variables, and between 1,200 and 1,600 clauses. We evaluate our methods with three different inner-optimization strategies.

\subsubsection{Baselines.}
We primarily compare our results against existing sub-QUBO methods that first convert a SAT instance to a QUBO one before decomposition into the QUBO space. These methods most closely follow the resource-constrained environment that minimizes the size of the QUBO matrix before optimization. 
For the decomposition in QUBO space, we use an energy and random-based approach. The energy selector chooses the QUBO variables that, when switched, increases the energy most (N.B., the energy function for a QUBO increases with satisfied clauses). We add a small amount of noise ($\mathcal{O}(1^{-6})$) to the energy values to introduce randomness when selecting variables with identical energy deltas. The random selector simply chooses $M$ arbitrary variables.

We also show results against three unconstrained solving methods. First, we convert a full SAT instance into a QUBO matrix and use the D-Wave implementation of Tabu search to optimize the matrix\footnote{https://docs.ocean.dwavesys.com/en/stable/docs\_hybrid/}. Second, we compare against Walk-SAT, a local heuristic that selects a random variable with probability $p$, otherwise chooses a variable that optimizes a specific function~\cite{selman1993local}. We use the following functions for Walk-SAT: energy: pick the variable which maximizes energy improvement; make: pick the variable which maximizes the number of clauses that go from satisfiable to unsatisfiable; and break: pick the variable which minimizes the number of clauses that go from unsatisfiable to satisfiable. Lastly, we use the exact solver \texttt{z3}\footnote{https://github.com/Z3Prover/z3}~\cite{de2008z3}. \texttt{z3} is more computationally intensive than our other unconstainted solvers, so we only include a subset of the results. However, since the solver is complete, this serves as ground truth. 

\subsubsection{Parameter Settings.}

We run experiments with $M = $ 10, 25, 50, 75, 100, 150, 200, 250, 500, 750, 100, 1500, 2000, and 2500.
When using Walk-SAT as an inner-optimizer, we use the break selection mechanism with $20 * M$ number of inner iterations. We use 100,000 iterations from the three Walk-SAT unconstrained solvers. All QUBO conversions, whether on a full or sub- SAT instance use an ``N + L'' conversion method~\cite{chancellor2016direct}. $max\_iters$ and $conv$ are set to 1,000 and 20, respectively. 
\section{Results}

Table~\ref{table:results} shows the results for the outer-loop selectors, inner-optimizers, and resource constraints on our nine randomly generated datasets. We also provide results for a sub-QUBO approach (i.e., one that decomposes in the QUBO space), as well as results from unconstrained solvers. We choose a select few values of $M$ for our tables but provide the complete results in the supplementary material. Our sub-SAT method often outperforms the sub-QUBO methods given the same hardware constraints, $M$, regardless of selector type. We note that when $M \ll N + L$, the energy selector for sub-QUBO performs poorly. Most QUBO variables have similar potential energy changes, so when $M$ is too small, the algorithm cannot explore the problem space enough. Unsurprisingly, the unconstrained solvers in the SAT space (Walk-SAT and z3) perform well, although the exact solver z3 did not finish on larger instances. Perhaps more surprisingly, our sub-SAT method outperforms a full QUBO solution on several different parameter configurations. 

We ran our framework on a subset of the crafted Max-Cut problems from the 2016 Max-SAT challenge. We evaluated all selectors with our sub-SAT approach. Full results appear in the supplementary materials. We use $M = 150$, $M = 50$, and $Q = 200$ for the Walk-Sat, z3, and QUBO + Tabu inner optimizers constraints. We found that the time to run z3 with $M = 100$ far exceeded the trials from the random dataset. We find this unsurprising as the hand-crafted instances are designed to be difficult for exact solvers. The QUBO + Tabu, z3, and Walk-SAT inner-optimizers achieve average energies of 232.22, 301.06, and 335.21, respectively.

\begin{table*}[t!]
    \centering
        \fontsize{9pt}{11pt}\selectfont
        \begin{tabular}{c c c | c c c | c c c | c c c} \specialrule{.15em}{.075em}{.075em}  
            & & & \multicolumn{3}{c |}{\textbf{100v}} & \multicolumn{3}{c |}{\textbf{200v}} & \multicolumn{3}{c}{\textbf{500v}} \\ 
            & & & \textbf{400c} & \textbf{425c} & \textbf{450c} & \textbf{800c} & \textbf{850c} & \textbf{900c} & \textbf{2000c} & \textbf{2125c} & \textbf{2250c} \\ \specialrule{.15em}{.075em}{.075em} 
            \textbf{Selector} & \textbf{Inner-Optimizer} & \textbf{M/Q} & \multicolumn{9}{c}{\textbf{Sub-SAT}} \\ \specialrule{.15em}{.075em}{.075em} 
            Energy & Walk-SAT & 75 & 0.30 & 0.83 & \textbf{1.63} & 2.19 & 3.15 & 5.08 & 8.43 & 11.81 & 16.39  \\
            Energy & Walk-SAT & 150 & & & & 0.27 & 1.33 & 3.24 & 4.81 & 8.29 & 12.44  \\
            Energy & Walk-SAT & 250 & & & & & & & 2.27 & 5.63 & \textbf{9.33}  \\ \hline
            Graph & Walk-SAT & 75 & 0.57 & 1.02 & 2.32 & 3.30 & 5.95 & 9.20 & 30.23 & 43.30 & 56.17  \\
            Graph & Walk-SAT & 150 & & & & 1.03 & 2.28 & 4.58 & 12.27 & 20.77 & 30.25  \\
            Graph & Walk-SAT & 250 & & & & & & & 4.95 & 11.78 & 19.65  \\ \hline
            Random & Walk-SAT & 75 & 0.54 & 1.26 & 2.71 & 9.49 & 14.17 & 19.88 & 70.39 & 87.00 & 107.21  \\
            Random & Walk-SAT & 150 & & & & 0.88 & 3.11 & 6.06 & 38.09 & 52.35 & 69.22  \\
            Random & Walk-SAT & 250 & & & & & & & 12.87 & 24.93 & 38.84  \\ \hline
            Softmax & Walk-SAT & 75 & \textbf{0.27} & \textbf{0.77} & 1.68 & 2.04 & 3.17 & 5.22 & 10.08 & 14.38 & 19.80  \\
            Softmax & Walk-SAT & 150 & & & & \textbf{0.29} & \textbf{1.28} & \textbf{3.03} & 4.87 & 8.28 & 14.01  \\
            Softmax & Walk-SAT & 250 & & & & & & & \textbf{2.18} & \textbf{5.32} & 10.20  \\ \specialrule{.15em}{.075em}{.075em} 
            Energy & z3 & 75 & \textbf{0.34} & \textbf{0.69} & 1.57 & 2.64 & 3.66 & 5.27 & 13.89 & 17.69 & 20.95  \\
            Energy & z3 & 100 & & & & \textbf{1.38} & 2.86 & \textbf{3.94} & 11.43 & 14.20 & 17.86  \\ \hline
            Graph & z3 & 75 & 0.61 & 1.13 & 2.04 & 2.26 & 3.88 & 5.28 & 11.60 & 18.14 & 24.43  \\
            Graph & z3 & 100 & & & & 1.77 & 2.97 & 5.13 & 8.93 & 14.01 & 18.69  \\ \hline
            Random & z3 & 75 & 0.57 & 0.96 & 1.79 & 3.04 & 4.84 & 6.99 & 29.88 & 40.61 & 52.12  \\
            Random & z3 & 100 & & & & 2.44 & 3.43 & 5.22 & 19.61 & 28.07 & 35.51  \\ \hline
            Softmax & z3 & 75 & 0.38 & 0.70 & 1.49 & 2.48 & 3.49 & 4.78 & 9.89 & 14.52 & 18.37  \\
            Softmax & z3 & 100 & & & & 1.58 & \textbf{2.67} & 4.19 & \textbf{8.32} & \textbf{11.39} & \textbf{15.66}  \\ 
            \specialrule{.15em}{.075em}{.075em} 
            Energy & QUBO + Tabu & 150 & \textbf{0.93} & 1.72 & \textbf{2.48} & \textbf{2.78} & \textbf{4.03} & 5.70 & \textbf{7.89} & 11.53 & 15.94  \\
            Energy & QUBO + Tabu & 200 & 1.36 & 1.89 & 2.89 & 3.10 & 4.69 & 5.96 & 8.01 & \textbf{10.78} & \textbf{15.88}  \\ \hline
            Graph & QUBO + Tabu & 150 & 1.33 & 2.11 & 3.13 & 4.01 & 9.68 & 8.28 & 15.00 & 30.19 & 34.74  \\
            Graph & QUBO + Tabu & 200 & 2.11 & 2.88 & 4.12 & 4.40 & 6.74 & 8.72 & 42.74 & 22.10 & 29.87  \\ \hline
            Random & QUBO + Tabu & 150 & 1.99 & 2.88 & 4.04 & 5.92 & 8.62 & 12.22 & 46.09 & 59.57 & 79.05  \\
            Random & QUBO + Tabu & 200 & 3.55 & 4.16 & 5.64 & 7.83 & 25.82 & 13.57 & 33.25 & 45.80 & 60.80  \\ \hline
            Softmax & QUBO + Tabu & 150 & 1.14 & \textbf{1.59} & 2.68 & 2.79 & 4.19 & \textbf{5.59} & 9.83 & 13.26 & 18.93  \\
            Softmax & QUBO + Tabu & 200 & 1.40 & 2.57 & 3.58 & 3.96 & 5.30 & 7.31 & 10.38 & 13.07 & 16.98  \\  \specialrule{.15em}{.075em}{.075em} 
            \textbf{Selector} & \textbf{Inner-Optimizer.} & \textbf{Q} & \multicolumn{9}{c}{\textbf{Sub-QUBO}} \\ \specialrule{.15em}{.075em}{.075em} 
            Energy & QUBO + Tabu & 250 & 5.94 & 7.69 & 9.38 & 26.37 & 34.37 & 42.36 & 190.29 & 195.94 & 204.13 \\
            Energy & QUBO + Tabu & 500 & & \textbf{1.66} & \textbf{2.56} & 11.12 & 14.37 & 17.97 & 182.53 & 201.04 & 218.24 \\ \hline
            Random & QUBO + Tabu & 250 & \textbf{4.25} & 5.59 & 7.54 & 15.70 & 18.75 & 22.54 & 55.60 & 65.71 & 76.03  \\
            Random & QUBO + Tabu & 500 & & 1.75 & 2.78 & \textbf{7.60} & \textbf{10.23} & \textbf{13.45} & \textbf{37.78} & \textbf{47.23} & \textbf{57.56}  \\  \specialrule{.15em}{.075em}{.075em} 
            \multicolumn{3}{c}{\textbf{Method}} & \multicolumn{9}{c}{\textbf{Unconstrained Solvers}} \\ \specialrule{.15em}{.075em}{.075em} 
            \multicolumn{3}{c}{QUBO + Tabu} & 1.70 & 2.33 & 3.67 & 4.68 & 11.50 & 11.56 & 38.47 & 32.55 & 57.22  \\
            \multicolumn{3}{c}{Walk-SAT Break} & \textbf{0.09} & \textbf{0.46} & \textbf{1.20} & \textbf{0.03} & \textbf{0.61} & 2.06 & \textbf{0.01} & \textbf{1.35} & 5.44  \\
            \multicolumn{3}{c}{Walk-SAT Energy} & 0.10 & \textbf{0.46} & 1.21 & 0.07 & 0.76 & \textbf{2.04} & 0.18 & 1.93 & \textbf{5.03}  \\
            \multicolumn{3}{c}{Walk-SAT Make} & 0.17 & 0.52 & 1.32 & 0.37 & 1.55 & 3.99 & 2.80 & 10.75 & 20.43  \\ \specialrule{.15em}{.075em}{.075em} 
            \multicolumn{3}{c}{z3} & 0.09 & 0.45 & 1.20\ & 0.01 & 0.58 & & & &  \\ \specialrule{.15em}{.075em}{.075em} 
        \end{tabular}
    \caption{Results for the sub-SAT decomposition algorithm compared to QUBO decomposition and full solvers. We outperform approcahes that decompose in the QUBO space, as well as a full QUBO solver. Blank spaces represent trials where the hardware constraint exceeds the problem size. We bold optimal values per column per section.}
    \label{table:results}
\end{table*}

\begin{figure}[t!]
    \centering
    \includegraphics[width=0.85\linewidth]{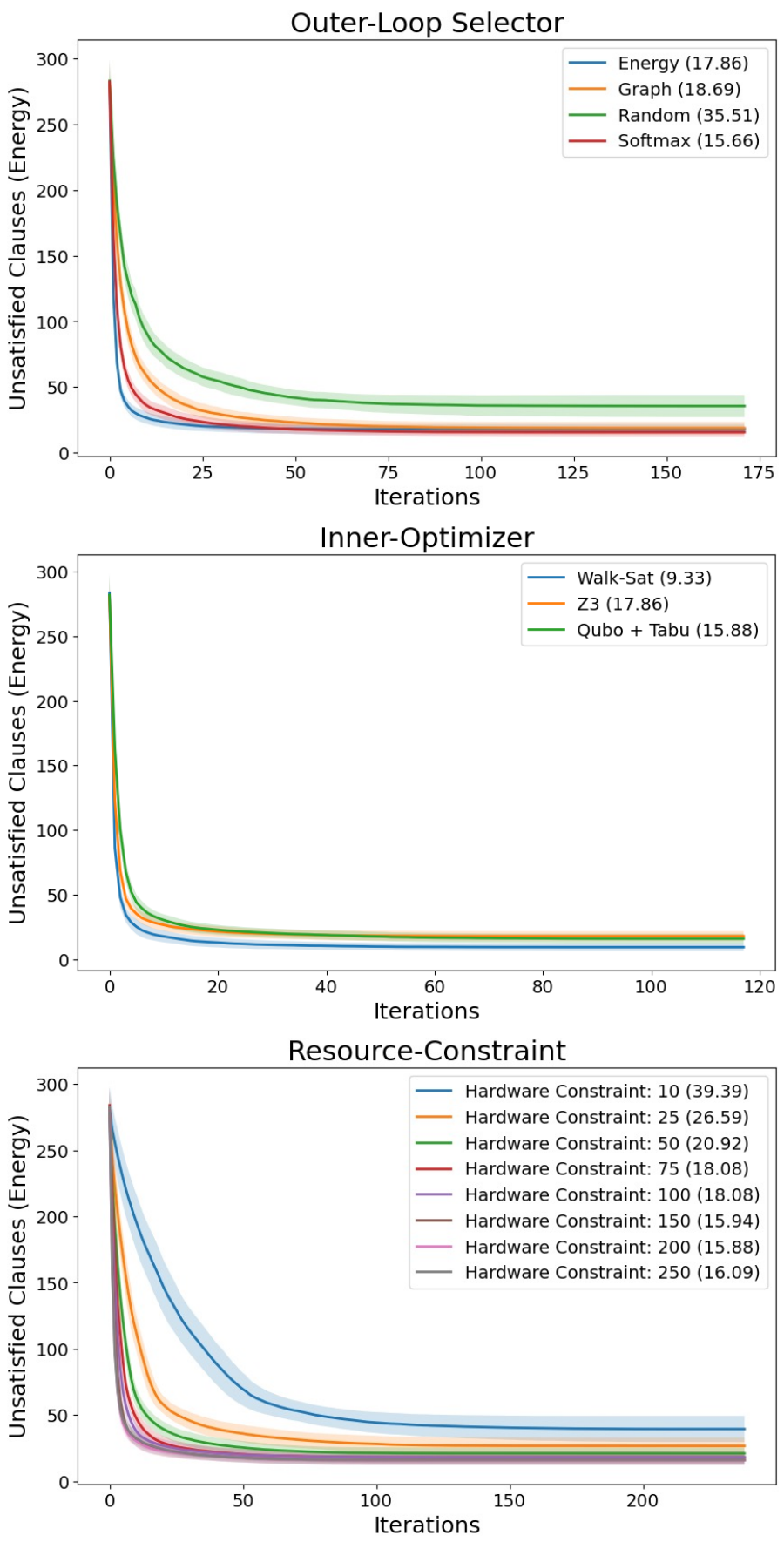}
    \caption{The energy and softmax selectors outperform the graph and random ones (top). Each of the inner-optimizers perform similarly well in their optimal configurations (middle). Our Max-SAT energies typically decrease as we reduce the constraints on the inner-optimizer (bottom).}
    \label{figure:ablation-study}
\end{figure}

In Figure~\ref{figure:ablation-study}, we present three plots on the largest dataset each for our various parameters: outer-loop selector, inner-optimizer, and hardware constraint. The top plot shows how both energy and softmax selectors tend to outperform in terms of Max-SAT. The middle plot compares the three inner optimizers, while the bottom plot shows how the performance changes as we increase the constraint $M$ (using z3 as the inner-optimizer). We see the expected behavior that we get increased Max-SAT performance with increasing $M$. We now go into a detailed comparison of the components of our approach:

\paragraph{Outer-Loop Selectors.}
The energy selector requires the fewest iterations of the inner-optimizers across all configurations (Table~\ref{table:complexity}). It converges sharply to the local minima before plateauing quickly (Figure~\ref{figure:ablation-study}, top). There is less variation between the graph, random, and softmax, in terms of iterations required. However, the computational complexity of energy and softmax far exceeds that of graph and random, leading to significantly more time required despite the fewer calls to the inner loop. Overall, the energy and softmax selectors produce better final energy levels than the random and graph-based approaches (Table~\ref{table:results}). In the three inner-optimizer configurations over the nine datasets, energy produces the best assignment fourteen times, softmax twelve times, and random once. 

\paragraph{Inner-Optimizers.}
Comparing the results between inner-optimizers is challenging since the hardware constraints carry different meanings in some instances. For example, with Walk-SAT as an optimizer, $M$ constrains the number of variables in the sub-SAT instance. In contrast with a QUBO conversion, $Q$ constrains the number of variables plus clauses in the sub-SAT instance. Thus, $M = 250$ allows for a significantly larger sub-problem of the CNF for Walk-SAT and z3 than for QUBO + Tabu. Furthermore, the computational costs of z3 make it intractable to use large values of $M$. However, we still show a comparison of the best methods using each inner-optimizer for our largest random datasest (Figure~\ref{figure:ablation-study}, middle). We see that Walk-SAT performs the best, followed by QUBO + Tabu, and then z3.

\subsubsection{Hardware Constraint.}
Often, but not always, increasing $M$ or $Q$ decreases the final energy state (Figure~\ref{figure:ablation-study}, bottom). The main exception to this rule has been when the inner-optimizer converts the sub-SAT problem into a QUBO one and optimizes in that space. We believe this is because optimizing large QUBO instances presents its own set of challenges. The SAT optimizers generally converge to good approximations relatively quickly, whereas the outputs from the QUBO inner-optimizer are more random, and less likely to represent a local minima in the SAT space. Table~\ref{table:results} shows results from a select few values of $M$ and $Q$. We provide a more comprehensive set of tables in the supplementary material. Empirically, we have found that with the QUBO inner-optimizer, values of $Q$ between 150 and 200 outperform others. Interestingly, we did not find this to be the case for the sub-QUBO methods. 

\subsection{Computational Costs}

We present the total clock time required for each of the selectors and inner-optimizers combinations in Table~\ref{table:complexity}. We accumulate results for $M$ and $Q$ sizes given a selector and inner-optimizer. Note, not every selector and optimizer will have the same number of solutions generated. For example, we constrain $M \leq 100$ for z3 because of its high cost. Thus, we have used z3 fewer times compared to Walk-SAT, as evidenced by the ``No. Runs'' column.

The energy and softmax selectors require significantly more computational overhead than the random and graph selectors. The overall cost is slightly reduced as the selectors tend to converge to their local minima at a faster rate. However, the reduction in calls to the inner-optimizer generally does not compensate for the expensive matrix multiplication required to identify the best $M$ candidates for selection. Compared to random, energy selectors require on average $1.66\times$ fewer iterations, but took $4.07\times$ more time. Softmax is even more intensive as the number of iterations is significantly more than the deterministic energy selector. Furthermore, as the hardware for the optimizers continues to improve and scale, the outer-loop begin to dominate the total running time, expanding the gap in performance between energy/softmax and graph/random.

\begin{table}[t!]
    \centering
    \fontsize{9pt}{11pt}\selectfont
    \setlength{\tabcolsep}{1mm}
    \begin{tabular}{c c c c c} \hline
        \textbf{Selector} & \textbf{Optimizer} & \textbf{No. Runs} & \textbf{Max. Iters} & \textbf{Time} \\ \hline 
        \multicolumn{5}{c}{\textbf{Sub-SAT}} \\ \hline 

        Energy & Walk-SAT & 5,400 & 102.65 & 8.51 d\\
        Graph & Walk-SAT & 5,400 & 126.94 & 3.45 d\\
        Random & Walk-SAT & 5,400 & 115.98 & 2.59 d\\
        Softmax & Walk-SAT & 5,400 & 128.00 & 10.67 d\\ \hline
        Energy & z3 & 4,200 & 81.86 & 3.36 d\\
        Graph & z3 & 4,200 & 135.57 & 0.67 d\\
        Random & z3 & 4,200 & 129.60 & 0.35 d\\
        Softmax & z3 & 4,200 & 121.55 & 6.01 d\\ \hline 
        Energy & QUBO + Tabu & 19,800 & 105.61 & 165.49 d\\
        Graph & QUBO + Tabu & 19,800 & 135.32 & 53.13 d\\
        Random & QUBO + Tabu & 19,800 & 126.09 & 27.11 d\\
        Softmax & QUBO + Tabu & 19,800 & 123.44 & 205.19 d\\ \hline 
        \multicolumn{5}{c}{\textbf{Sub-QUBO}} \\ \hline 
        Energy & QUBO + Tabu & 19,800 & 63.29 & 172.44 d\\
        Random & QUBO + Tabu & 19,800 & 214.71 & 55.96 d\\ \hline
        \multicolumn{5}{c}{\textbf{Unconstrained Solvers}} \\ \hline 
        \multicolumn{2}{c}{QUBO + Tabu} & 900 & 115.56 & 0.31 d\\
        \multicolumn{2}{c}{Walk-SAT Break} & 900 & 100,000 & 0.40 d\\
        \multicolumn{2}{c}{Walk-SAT Energy} & 900 & 100,000 & 0.21 d\\
        \multicolumn{2}{c}{Walk-SAT Make} & 900 & 100,000 & 0.26 d\\
        \multicolumn{2}{c}{z3} & 500 & 1.00 & 0.33 d\\
        \hline
    \end{tabular}
    \caption{The energy and softmax selectors add a significant amount of overhead to the overall running time.}
    \label{table:complexity}
\end{table}
\section{Conclusions}

We propose a resource-constrained heuristic for Max-SAT problems that decomposes the problem in the SAT space before calling a optimized solver. We decouple the problem partitioning and sub-problem solving to enable greater flexibility. We contrast with existing solutions that first reformulate a SAT instance into a QUBO problem before the decomposition step. We outperform such existing strategies on randomly generated instances of various variable and clause size. Our method enables us to make sure of  specialized hardware, such as quantum annealers, that offer significant throughput when presented with small QUBO instances. We demonstrate our results on a thousands of random and crafted Max-SAT instances.

\section{Acknowledgments}
This material is based upon work supported by the Defense Advanced Research Projects Agency (DARPA) through the United States Air Force Research Laboratory (ARFL) Contract No. FA8750-23-C-1001. The views, opinions, and/or findings expressed are those of the author(s) and should not be interpreted as representing the official views or policies of the Department of Defense or the U.S. Government.

\bibliography{references}

\begin{thebibliography}{27}
\providecommand{\natexlab}[1]{#1}

\bibitem[{Alasow, Jin, and Perkowski(2022)}]{alasow2022quantum}
Alasow, A.; Jin, P.; and Perkowski, M. 2022.
\newblock Quantum Algorithm for Variant Maximum Satisfiability.
\newblock \emph{Entropy}, 24(11): 1615.

\bibitem[{Ans{\'o}tegui, Gir{\'a}ldez-Cru, and Levy(2012)}]{ansotegui2012community}
Ans{\'o}tegui, C.; Gir{\'a}ldez-Cru, J.; and Levy, J. 2012.
\newblock The community structure of SAT formulas.
\newblock In \emph{International Conference on Theory and Applications of Satisfiability Testing}, 410--423. Springer.

\bibitem[{Argelich and Many{\`a}(2006)}]{argelich2006exact}
Argelich, J.; and Many{\`a}, F. 2006.
\newblock Exact Max-SAT solvers for over-constrained problems.
\newblock \emph{Journal of Heuristics}, 12(4): 375--392.

\bibitem[{As{\'\i}n~Ach{\'a} and Nieuwenhuis(2014)}]{asin2014curriculum}
As{\'\i}n~Ach{\'a}, R.; and Nieuwenhuis, R. 2014.
\newblock Curriculum-based course timetabling with SAT and MaxSAT.
\newblock \emph{Annals of Operations Research}, 218: 71--91.

\bibitem[{Booth, Reinhardt, and Roy(2017)}]{booth2017partitioning}
Booth, M.; Reinhardt, S.~P.; and Roy, A. 2017.
\newblock Partitioning optimization problems for hybrid classical.
\newblock \emph{quantum execution. Technical Report}, 01--09.

\bibitem[{Chancellor et~al.(2016)Chancellor, Zohren, Warburton, Benjamin, and Roberts}]{chancellor2016direct}
Chancellor, N.; Zohren, S.; Warburton, P.~A.; Benjamin, S.~C.; and Roberts, S. 2016.
\newblock A direct mapping of max k-SAT and high order parity checks to a chimera graph.
\newblock \emph{Scientific reports}, 6(1): 37107.

\bibitem[{Date et~al.(2019)Date, Patton, Schuman, and Potok}]{date2019efficiently}
Date, P.; Patton, R.; Schuman, C.; and Potok, T. 2019.
\newblock Efficiently embedding QUBO problems on adiabatic quantum computers.
\newblock \emph{Quantum Information Processing}, 18: 1--31.

\bibitem[{De~Moura and Bj{\o}rner(2008)}]{de2008z3}
De~Moura, L.; and Bj{\o}rner, N. 2008.
\newblock Z3: An efficient SMT solver.
\newblock In \emph{International conference on Tools and Algorithms for the Construction and Analysis of Systems}, 337--340. Springer.

\bibitem[{Fu and Malik(2006)}]{fu2006solving}
Fu, Z.; and Malik, S. 2006.
\newblock On solving the partial MAX-SAT problem.
\newblock In \emph{International Conference on Theory and Applications of Satisfiability Testing}, 252--265. Springer.

\bibitem[{Hickey and Bacchus(2022)}]{hickey2022large}
Hickey, R.; and Bacchus, F. 2022.
\newblock Large Neighbourhood Search for Anytime MaxSAT Solving.
\newblock In \emph{IJCAI}, 1818--1824.

\bibitem[{Hoos and St{\"u}tzle(2000)}]{hoos2000local}
Hoos, H.~H.; and St{\"u}tzle, T. 2000.
\newblock Local search algorithms for SAT: An empirical evaluation.
\newblock \emph{Journal of Automated Reasoning}, 24(4): 421--481.

\bibitem[{Kalinin et~al.(2023)Kalinin, Mourgias-Alexandris, Ballani, Berloff, Clegg, Cletheroe, Gkantsidis, Haller, Lyutsarev, Parmigiani et~al.}]{kalinin2023analog}
Kalinin, K.~P.; Mourgias-Alexandris, G.; Ballani, H.; Berloff, N.~G.; Clegg, J.~H.; Cletheroe, D.; Gkantsidis, C.; Haller, I.; Lyutsarev, V.; Parmigiani, F.; et~al. 2023.
\newblock Analog Iterative Machine (AIM): using light to solve quadratic optimization problems with mixed variables.
\newblock \emph{arXiv preprint arXiv:2304.12594}.

\bibitem[{Karloff and Zwick(1997)}]{karloff19977}
Karloff, H.; and Zwick, U. 1997.
\newblock A 7/8-approximation algorithm for MAX 3SAT?
\newblock In \emph{Proceedings 38th Annual Symposium on Foundations of Computer Science}, 406--415. IEEE.

\bibitem[{Krentel(1986)}]{krentel1986complexity}
Krentel, M.~W. 1986.
\newblock The complexity of optimization problems.
\newblock In \emph{Proceedings of the eighteenth annual ACM symposium on Theory of computing}, 69--76.

\bibitem[{Lauria et~al.(2017)Lauria, Elffers, Nordstr{\"o}m, and Vinyals}]{lauria2017cnfgen}
Lauria, M.; Elffers, J.; Nordstr{\"o}m, J.; and Vinyals, M. 2017.
\newblock CNFgen: A generator of crafted benchmarks.
\newblock In \emph{Theory and Applications of Satisfiability Testing--SAT 2017: 20th International Conference, Melbourne, VIC, Australia, August 28--September 1, 2017, Proceedings 20}, 464--473. Springer.

\bibitem[{Moln{\'a}r et~al.(2018)Moln{\'a}r, Moln{\'a}r, Varga, Toroczkai, and Ercsey-Ravasz}]{molnar2018continuous}
Moln{\'a}r, B.; Moln{\'a}r, F.; Varga, M.; Toroczkai, Z.; and Ercsey-Ravasz, M. 2018.
\newblock A continuous-time MaxSAT solver with high analog performance.
\newblock \emph{Nature communications}, 9(1): 4864.

\bibitem[{Morgado et~al.(2013)Morgado, Heras, Liffiton, Planes, and Marques-Silva}]{morgado2013iterative}
Morgado, A.; Heras, F.; Liffiton, M.; Planes, J.; and Marques-Silva, J. 2013.
\newblock Iterative and core-guided MaxSAT solving: A survey and assessment.
\newblock \emph{Constraints}, 18: 478--534.

\bibitem[{N{\"u}{\ss}lein et~al.(2023)N{\"u}{\ss}lein, Roch, Gabor, Stein, Linnhoff-Popien, and Feld}]{nusslein2023black}
N{\"u}{\ss}lein, J.; Roch, C.; Gabor, T.; Stein, J.; Linnhoff-Popien, C.; and Feld, S. 2023.
\newblock Black box optimization using QUBO and the cross entropy method.
\newblock In \emph{International Conference on Computational Science}, 48--55. Springer.

\bibitem[{Orvalho, Manquinho, and Martins(2023)}]{orvalho2023}
Orvalho, P.; Manquinho, V.; and Martins, R. 2023.
\newblock UpMax: User partitioning for MaxSAT.
\newblock In \emph{26th International Conference on Theory and Applications of Satisfiability Testing (SAT 2023)}. Schloss Dagstuhl-Leibniz-Zentrum f{\"u}r Informatik.

\bibitem[{Palubeckis(2004)}]{palubeckis2004multistart}
Palubeckis, G. 2004.
\newblock Multistart tabu search strategies for the unconstrained binary quadratic optimization problem.
\newblock \emph{Annals of Operations Research}, 131: 259--282.

\bibitem[{Selman, Kautz, and Cohen(1993)}]{selman1993local}
Selman, B.; Kautz, H.~A.; and Cohen, B. 1993.
\newblock Local search strategies for satisfiability testing.
\newblock \emph{Cliques, coloring, and satisfiability}, 26: 521--532.

\bibitem[{Selman, Mitchell, and Leveque(1992)}]{selman1992new}
Selman, B.; Mitchell, D.; and Leveque, H. 1992.
\newblock A new method for solving hard satisfiability problems.
\newblock In \emph{Proceedings of the tenth national conference on artificial intelligence (AAAI-92)}, 440--446.

\bibitem[{Shaw(1998)}]{shaw1998using}
Shaw, P. 1998.
\newblock Using constraint programming and local search methods to solve vehicle routing problems.
\newblock In \emph{International conference on principles and practice of constraint programming}, 417--431. Springer.

\bibitem[{Sinjorgo and Sotirov(2023)}]{sinjorgo2023solving}
Sinjorgo, L.; and Sotirov, R. 2023.
\newblock On Solving MAX-SAT Using Sum of Squares.
\newblock \emph{INFORMS Journal on Computing}.

\bibitem[{Spears(1993)}]{spears1993simulated}
Spears, W.~M. 1993.
\newblock Simulated annealing for hard satisfiability problems.
\newblock \emph{Cliques, Coloring, and Satisfiability}, 26: 533--558.

\bibitem[{Xing and Zhang(2005)}]{xing2005maxsolver}
Xing, Z.; and Zhang, W. 2005.
\newblock MaxSolver: An efficient exact algorithm for (weighted) maximum satisfiability.
\newblock \emph{Artificial intelligence}, 164(1-2): 47--80.

\bibitem[{Zaman, Tanahashi, and Tanaka(2021)}]{zaman2021pyqubo}
Zaman, M.; Tanahashi, K.; and Tanaka, S. 2021.
\newblock PyQUBO: Python library for mapping combinatorial optimization problems to QUBO form.
\newblock \emph{IEEE Transactions on Computers}, 71(4): 838--850.

\end{thebibliography}

\end{document}